\documentclass{l4dc2025}

\title[Task-Parameter Nexus]{Task-Parameter Nexus: Task-Specific Parameter Learning for Model-Based Control}
\usepackage{times}

\coltauthor{\Name{Sheng Cheng} \Email{chengs@illinois.edu}\\
 \Name{Ran Tao} \Email{rant3@illinois.edu}\\
 \Name{Yuliang Gu} \Email{yuliang3@illinois.edu}\\
 \Name{Shenlong Wang} \Email{shenlong@illinois.edu}\\
 \Name{Xiaofeng Wang} \Email{wangxi@cec.sc.edu}\\
 \Name{Naira Hovakimyan} \Email{nhovakim@illinois.edu}
 }

\usepackage{graphics} 
\usepackage{times} 
\usepackage{amsmath} 
\usepackage{amssymb}  
\usepackage{xcolor}
\usepackage{bm}
\usepackage{cancel}
\usepackage{color}
\usepackage{setspace}
\usepackage{wrapfig}
\usepackage{tikz}
\usetikzlibrary{shapes,arrows.meta,calc}
\usetikzlibrary{positioning}
\usepackage{caption}
\usepackage{subcaption}
\usepackage{booktabs}
\usepackage{multirow}
\usepackage{multicol}
\usepackage{soul}
\usepackage{cite}
\usepackage[normalem]{ulem}
\usepackage{xcolor}
\usepackage{mathtools}
\usepackage{stfloats}

\usepackage{enumitem}
\usepackage{courier}
\usepackage{algorithm}
\usepackage{algorithmic}
\usepackage{makecell}

\usepackage{pifont}
\usepackage{url}

\newcommand{\todocite}[1]{\textcolor{red}{[TODO(cite)]}}

\allowdisplaybreaks

\newcommand{\norm}[1]{\lVert #1 \rVert}

\begin{document}

\maketitle

\vspace{-1cm}

\begin{abstract}
This paper presents the Task-Parameter Nexus (TPN), a learning-based approach for online determination of the (near-)optimal control parameters of model-based controllers (MBCs) for tracking tasks. In TPN, a deep neural network is introduced to predict the control parameters for any given tracking task at runtime, especially when optimal parameters for new tasks are not immediately available. 
To train this network, we constructed a trajectory bank with various speeds and curvatures that represent different motion characteristics. Then, for each trajectory in the bank, we auto-tune the optimal control parameters offline and use them as the corresponding ground truth.  With this dataset, the TPN is trained by supervised learning. We evaluated the TPN on the quadrotor platform.
In simulation experiments, it is shown that the TPN can predict near-optimal control parameters for a spectrum of tracking tasks, demonstrating its robust generalization capabilities to unseen tasks.
\end{abstract}



\section{Introduction}

A typical robot control problem often starts with a specific task (like manipulators operating in warehouses for moving objects around, or aerial robots transporting packages), for which a suitable control solution will then be designed to accomplish this task. When the task changes, one may have to re-tune the control parameters; otherwise, the original parameters might not be able to offer a satisfactory performance and even jeopardize task completion. 
For instance, consider a PD-type controller for a quadrotor's translational control. One set of control parameters cannot cater to different tasks uniformly well: if the task requires the quadrotor to hover, then the D-gain in the PD controller needs to be dominant to allow sufficient damping so that the quadrotor can smoothly return to the hover position when it is perturbed. On the contrary, if the task needs the quadrotor to track an aggressive trajectory (with high speeds or sharp turns), the D-gain needs to be reduced and the P-gain might be raised at the same time for sufficient agility and responsiveness to the fast-changing reference. 
A traditional way to address this issue is to use gain-scheduling~\citep{rugh2000research}, which, however,  cannot handle tasks that are not predefined.  These observations lead to a challenging question: given a task that could be new and not predefined, how can we online determine the associated optimal or near-optimal control parameters in a given model-based control structure?

\begin{wrapfigure}{r}{0.56\textwidth}
      \vspace{-1mm}
     \includegraphics[width=0.56\textwidth]{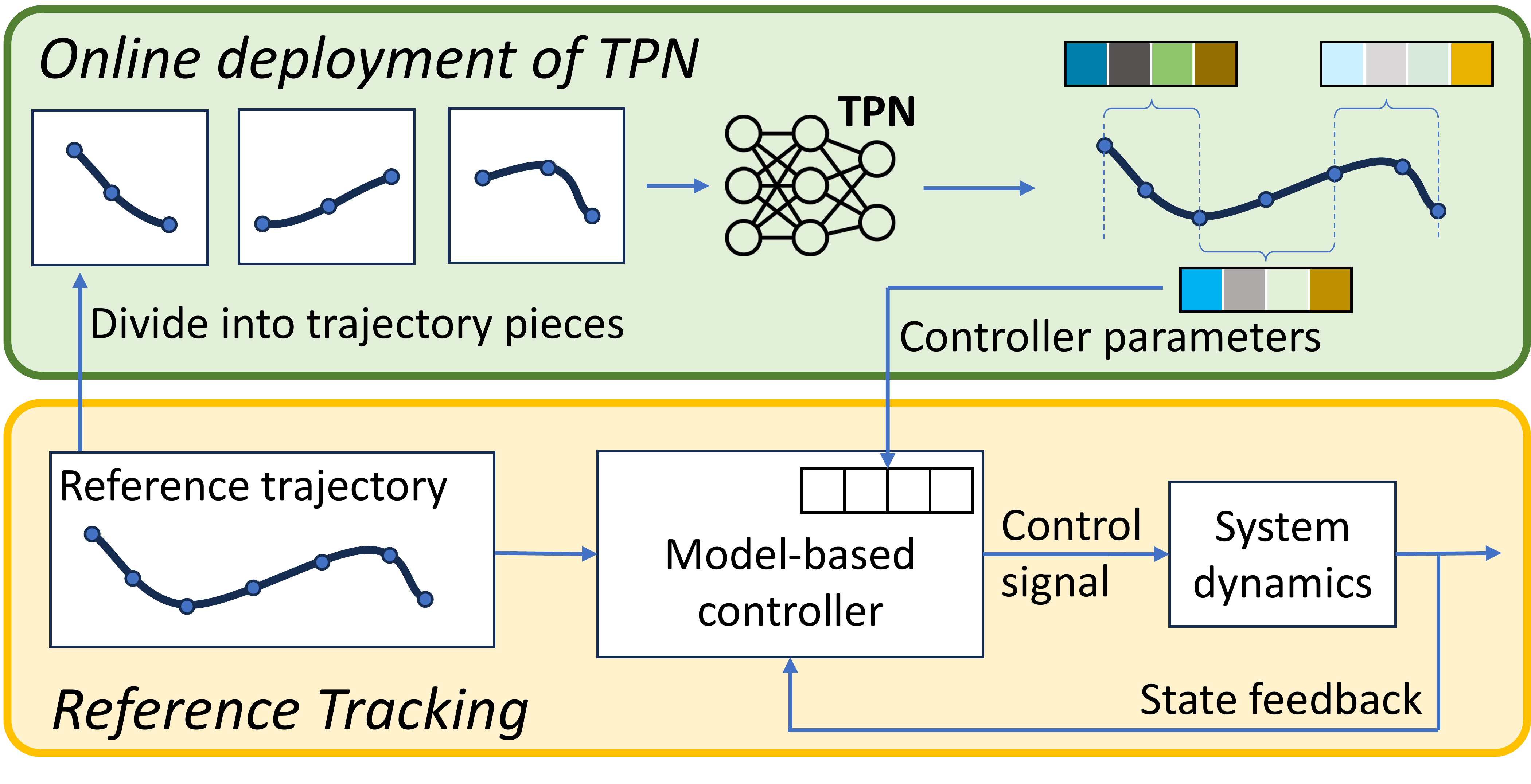}
     \caption{An illustration of the Task-Parameter Nexus when applied to a model-based controller.}
     \label{fig: TPN illustration}
     \vspace{-3mm}
\end{wrapfigure}

This paper introduces such a tool that enables the system to learn the control parameters when operating different tasks. For the purposes of illustration, we consider a quadrotor platform where a \textit{task} is defined as tracking a specific trajectory.
Different tasks imply tracking different trajectories with distinct motion characteristics. Since each trajectory is potentially associated with a set of optimal control parameters, there exists a nontrivial mapping from tasks to control parameters.  We approximate this mapping using a deep neural network (DNN), called Task-Parameter Nexus (TPN). Given a task characterized by a reference trajectory, the TPN predicts the (near-) optimal control parameters of a model-based controller (MBC) for tracking purposes, as shown in Fig.~\ref{fig: TPN illustration}.  To train the TPN, we first create a trajectory bank that includes trajectories with various motion characteristics categorized by speed and curvature.  Note that varying speeds and curvatures present distinct challenges for tracking translational and rotational motions, respectively.
By changing the speed and curvature, the trajectory bank covers rich enough tracking tasks that the quadrotor may encounter. Consequently, for each trajectory from the bank, we obtain the optimal control parameters, also referred to as ``expert parameters,'' by performing auto-tuning~\citep{cheng2022difftune} offline. We use them as the ground-truth labels.  With the pairs of trajectories in the bank and corresponding expert parameters, we train the TPN in a supervised manner. Leveraging the generalization capability of neural networks,  the TPN can possibly infer appropriate parameters for unseen testing tasks, making the TPN suitable for adapting the system to new tasks for which expert parameters are not immediately available. In simulation experiments, the TPN demonstrates robust generalization capabilities across a spectrum of testing tasks, predicting near-optimal control parameters.

The contributions of this paper are summarized as follows: (i) We present the TPN, an inference tool to determine appropriate control parameters for a system to perform new/unseen tasks when suitable parameters are not immediately available during operation. (ii) We construct a trajectory bank, from which we sample tasks and use auto-tuning to obtain optimal control parameters as the labels to train the TPN.  The trajectory bank contains trajectories systematically spanning a wide range of speeds and curvatures, which can be used for training policies or benchmarking different control laws/policies. (iii) We introduce a batch version of DiffTune~\citep{cheng2022difftune} as the controller auto-tuning scheme to obtain the expert parameters, which extends the original single-task scenario to a multi-task scenario. The batch-DiffTune avoids overfitting the parameters to one specific task and improves the robustness of the auto-tuned parameters.


\section{Related Literature}\label{sec: literature review}

\paragraph{Limited Generalization Capability of Low-Level Control in Robotics}
Large language models~\citep{touvron2023llama,Mercat2024Linearizing} and vision language models~\citep{wang2024visionllm,liu2023improved,lu2024deepseekvl} have achieved remarkable success in understanding and generating human language, recognizing objects, and interpreting scenes. 
This progress is largely due to the availability of large-scale datasets and the ability to learn patterns from vast amounts of visual and textual information, enabling these models to generalize well across different contexts and scenarios~\citep{brohan2023rt}. However, limited generalization capability in low-level control has been reported~\citep{brohan2023rt} due to relatively scarce data available for a model to capture unseen motions. Furthermore, even if a large foundation model generalizes effectively at the motion level, the low-level control, which often follows the model-based design and requires task-specific tuning~\citep{hwangbo2019learning,bohez2022imitate}, may not generalize equally well. In this work, we address the problem of limited generalization capability of the low-level control by inferring control parameters for achieving near-optimal performance over unseen control tasks.

For model-based controllers, one can obtain optimal control performance by auto-tuning the controller subject to known tasks. Many auto-tuning schemes have been proposed in recent years that leverage learning-enabled approaches, e.g., Gaussian Processes~\citep{berkenkamp2016safe,brunke2022safe,GIBO} and sampling~\citep{loquercio2022autotune,song2022policy}. Autodifferentiation has also been used for auto-tuning~\citep{cheng2022difftune,cheng2023difftuneplus,tao2023DT-MPC,kumar2021diffloop} when the system dynamics and controller are differentiable. However, the above-introduced auto-tuning schemes usually apply to a single-task setting, leading to likely overfitting for a particular task, thus limiting the generalization capability to other tasks. In this work, we extend the single-task tuning scheme~\citep{cheng2022difftune} to a batch setting, which enhances the robustness of the tuned parameters to spatial variations of the tasks.

\paragraph{Quadrotor Benchmark Trajectories} Previous works~\citep{sun2022comparative,saviolo2022physics,wu20221,wu2023L1Quad} focusing on quadrotor control design often included a set of trajectories for benchmarking the control laws or policies. These trajectories are mostly trigonometric curves, for which the trajectories can take the shapes of the lemniscate, circle, 
or parabola by choosing different trigonometric functions on different axes in a coordinated frame~\citep{saviolo2022physics}. 
To quantify the motion characteristics of trajectories, people use the maximum speed or acceleration as the index~\citep{torrente2021data}. 
By stretching the top linear speed in a range, the challenges of translational motion tracking are well characterized. However, these trajectories lack independent control over the characteristics for the \textit{rotational} motions because indices for rotational motions, such as maximum angular velocity or acceleration, depend on the stretched maximum linear speeds subject to a particular trajectory with a fixed geometry. In contrast, we provide a trajectory bank that yields independently controlled motion characteristics in terms of speed and curvature, which provides a dataset covering a wide range of trajectories that can generalize to unseen trajectories associated with new tasks.

\section{Overview of the Task-Parameter Nexus}\label{sec: TPN overview}
\noindent
\textbf{Notations:} We denote by $\mathbb R^n$ the $n$-dimensional real vector space, by $\mathbb R^+$ the set of the positive real numbers, and $\mathbb R^+_0=\mathbb R^+\cup \{0\}$. Given $\delta\in \mathbb R^+_0$, $\lfloor \delta \rfloor$ is the largest integer that is less than or equal to $\delta$ and $\lceil \delta \rceil$ is the smallest integer that is greater than or equal to $\delta$.  The terms ``task'' and ``reference trajectory'' are interchangeable in the following discussions.

Consider a discrete-time dynamical system
\begin{equation}\label{eq: dynamics}
    \boldsymbol{x}_{k+1} = f(\boldsymbol{x}_k,\boldsymbol{u}_k), \quad
    \boldsymbol{y}_k =g(\boldsymbol{x}_k),
\end{equation}
where $\boldsymbol{x}_k \in \mathbb R^{n_{\boldsymbol{x}}}$, $\boldsymbol{u}_k \in \mathbb R^{n_{\boldsymbol{u}}} $, and $\boldsymbol{y}_k \in \mathbb R^{n_{\boldsymbol{y}}}$ are the state, control input, and output, respectively, with appropriate dimensions, the initial state $\boldsymbol{x}_0$ is known, and $f,g$ are known functions. Let $t_s$ be the sampling period, based on which \eqref{eq: dynamics} is obtained from a continuous-time model. Let $\{\bar{\boldsymbol{y}}_{k}\}_{k=1}^N$ define the reference output trajectory (also known as ``task'') for $\boldsymbol{y}_k$ to track and 
\begin{equation}
    \boldsymbol{u}_k = h(\boldsymbol{x}_{0:k},\bar{\boldsymbol{y}}_{1:k+1},\boldsymbol{\theta}) ,~~~ k \in \{0,1,\cdots, N-1\}\label{eq: feedback controller}
\end{equation}
be a feedback controller for achieving it,
where $N$ is the number of steps under consideration, $\boldsymbol{x}_{0:k} = \{\boldsymbol{x}_i\}_{i=0}^k$, $\bar{\boldsymbol{y}}_{1:k+1}  = \{\bar {\boldsymbol{y}}_i\}_{i=1}^{k+1}$, $h$ refers to the known model-based control law,
$\boldsymbol{\theta} \in \Theta$ denotes the control parameters, and $\Theta$ represents the feasible set of parameters for which the system's stability can be guaranteed either analytically or empirically. Since $h$ is pre-defined, the tracking cost is uniquely determined by the value of $\boldsymbol{\theta}$ and therefore can be defined as $L_N(\boldsymbol{\theta};\bar{\boldsymbol{y}}_{1:N})$, where the subscript $N$ represents the cost over an $N$-step horizon.


This paper addresses the problem of how to predict the near-optimal control parameters at runtime for tracking an arbitrarily given trajectory. We assume that there exists a nontrivial mapping from the tasks to the optimal control parameters for a given controller. We attempted to approximate this mapping via a DNN and name the resulting network as Task-Parameter Nexus (TPN), which is formally defined as follows: 
Let $\mathcal{T}_M$ denote the set of all reference trajectories with a length of $M$ steps. For an arbitrary 
$T \in \mathcal{T}_M$, the TPN $\phi: \ \mathcal{T}_M \rightarrow {\Theta}$ maps the task $T$ 
to the parameter $\boldsymbol{\theta} \in \Theta$, i.e., $\boldsymbol{\theta}  = \phi(T)$,  which will then be used in the controller $h$ for the tracking task $T$. 
Note that the entire horizon of a task, $N$, is usually greater than $M$.  In that case, the reference trajectory $\{\bar{\boldsymbol{y}}_{k}\}_{k=1}^N$ will be split into consecutive pieces where each piece includes $M$ steps. Here we assume that $N$ is a multiple of $M$; otherwise, we can add $\lceil \frac{N}{M} \rceil M - N$ number of $\bar{\boldsymbol{y}}_N$ at the end of the reference trajectory to make it a multiple of $M$. For each trajectory piece, the TPN will predict a corresponding parameter $\boldsymbol{\theta}$. As a result, the value of $\boldsymbol{\theta}$ will be updated every $M$ steps. Formally, let task $T_i = \{\bar{\boldsymbol{y}}_j\}_{j=iM+1}^{(i+1)M},~i \in \{0, 1,\cdots,\frac{N}{M}-1\}$ and $\boldsymbol{\theta}_i  = \phi(T_i)$. The control law in~\eqref{eq: feedback controller} becomes:
\begin{equation*}
u_k =h\left(\boldsymbol{x}_{0:k},\bar{\boldsymbol{y}}_{1:k+1}, \boldsymbol{\theta}_{\lfloor \frac{k}{M} \rfloor}\right) =h\left(\boldsymbol{x}_{0:k},\bar{\boldsymbol{y}}_{1:k+1},\phi\left(T_{\lfloor \frac{k}{M} \rfloor}\right)\right),~~~ k  \in \{0,1,\cdots, N-1\},
\end{equation*}
as illustrated in Fig.~\ref{fig: TPN illustration}. 

The question now is how to train the TPN $\phi$: we need a training dataset, which will cover a sufficient range of tasks that the system is anticipated to perform, as well as a suitable DNN.  We will discuss the development of these two components in the following section, using a quadrotor platform for illustration.

\section{Case Study: TPN for Agile Trajectory Tracking Tasks of Quadrotors}
\label{sec: case study}
In this section, we will introduce the TPN as well as the training dataset, using trajectory tracking by a quadrotor as a case study. 
We will first consider how to generate a bank of $M$-step tasks/trajectories as the training tasks for the TPN. Subsequently, we will introduce the batch-DiffTune method to obtain the expert parameters for each task and use them as the ground truth.  With the task-parameter pairs in the training set, we can train the TPN using supervised learning.

\begin{figure}[t]
    \centering
    \includegraphics[width = \textwidth]{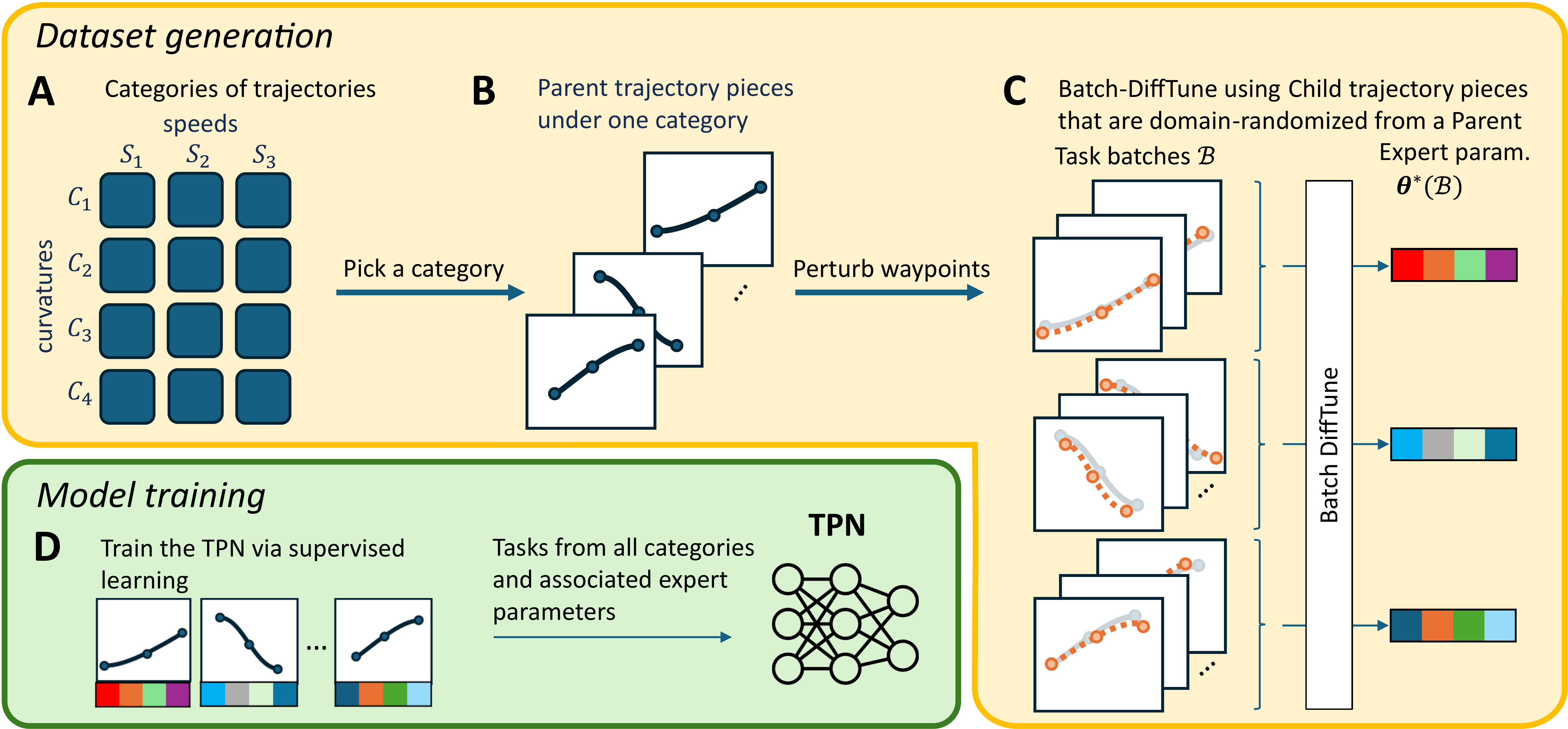}
    \caption{Illustration of the trajectory bank (consisting of trajectories under the hierarchy of category-parent-child-piece) and task batches for batch-DiffTune. Each trajectory piece is a \textit{task}.}
    \label{fig: generation of trajectory bank}

    \vspace{-4mm}
\end{figure}

\subsection{Generating the Bank of Reference Trajectories} \label{sec: task generation}

We seek to construct a trajectory bank that covers a wide range of tasks under two motion characteristics of a trajectory: speed and curvature. Speed characterizes translational aggressiveness, where higher speeds demand quicker control responses. Furthermore, as the speed increases, the aerodynamic effects (e.g., drag) become more significant, which challenges the translational control of the quadrotor. Curvature characterizes rotational aggressiveness, where high curvature requires drastic changes in attitude. The quadrotor must rapidly rotate to align with the reference trajectory, challenging the attitude control system. 

We will use speed and curvature as the coordinates to jointly span a range of motion characteristics. Specifically, let $\mathcal{S} \subset \mathbb R^+_0$ and $\mathcal{C} \subset \mathbb R^+_0$ be the intervals of interest on speed and curvature, respectively.  Then we partition the set $\mathcal{S}$ into $n_{\mathcal{S}}$ sub-intervals, denoted as $\{\mathcal{S}_i\}_{i=1}^{n_{\mathcal{S}}}$, such that $\mathcal{S} = \cup_{i=1}^{n_{\mathcal{S}}} \mathcal{S}_i$ and $\mathcal{S}_i \cap \mathcal{S}_j = \emptyset$ for any $i \not = j$. Similarly, we can define $\mathcal C_i \in \mathcal C$ such that $\mathcal{C} = \cup_{i=1}^{n_{\mathcal{C}}} \mathcal{C}_i$ and $\mathcal{C}_i \cap \mathcal{C}_j = \emptyset$ for any $i \not = j$, where $n_{\mathcal{C}}$ is the number of sub-intervals for curvature.
Thus, a combination of speed and curvature sub-intervals, $\mathcal{S}_i\mathcal{C}_j$, defines a category of trajectories that share similar motion characteristics with speed in $\mathcal{S}_i$ and curvature in $\mathcal{C}_j$.
Therefore, the trajectory bank will be the collection of all trajectories in all categories, i.e., $\cup_{i=1}^{n_{\mathcal S}} \cup_{j = 1}^{n_{\mathcal C}} \mathcal{S}_i \mathcal{C}_j$.  

To generate the trajectories in $\mathcal{S}_i \mathcal{C}_j$, we first need to produce a sequence of waypoints $\{\boldsymbol{p}_d\}_{d = 1}^D$ and the associated time of arrival (ToA) $\{t_d\}_{d = 1}^D$ at these waypoints that fulfill the requirement of speed $\mathcal S_i$ and curvature $\mathcal C_j$.
Without loss of generality, we fix the time increment $\Delta t$ ($\gg t_s$, the sampling period) between adjacent ToAs, i.e., $t_{d+1} - t_d = \Delta t$ for any $d \in \{1,2,\dots, D-1\}$.
Starting from $\boldsymbol{p}_1$, we iteratively generate a list of waypoints, satisfying two requirements:

\begin{minipage}{.45\linewidth}
\begin{equation} \label{eq: march speed requirement}
  \norm{\boldsymbol{p}_{d+1} - \boldsymbol{p}_d} = v_i \Delta t,
\end{equation}
\end{minipage}%
\begin{minipage}{.45\linewidth}
\begin{equation}\label{eq: curvature requirement}
 \frac{1}{R(\boldsymbol{p}_{d}, \boldsymbol{p}_{d+1}, \boldsymbol{p}_{d+2})} \in \mathcal{C}_j,
\end{equation}
\end{minipage}

\noindent where $v_i$ is the median speed of $\mathcal{S}_i$ and $1/R(\boldsymbol{p}_{d}, \boldsymbol{p}_{d+1}, \boldsymbol{p}_{d+2})$ is the Menger curvature~\citep{menger1930untersuchungen}, defined as the reciprocal of the radius of a circle that sequentially passes through the three waypoints $\boldsymbol{p}_{d}$, $\boldsymbol{p}_{d+1}$, and $\boldsymbol{p}_{d+2}$. Equation~\eqref{eq: march speed requirement} controls the marching speed along the generated trajectory around the designated speed $v_i$ of $\mathcal{S}_i$, whereas \eqref{eq: curvature requirement} enforces the curvature range to be within $\mathcal{C}_j$. Note that given $\boldsymbol{p}_{d}$, the waypoints $\boldsymbol{p}_{d+1}$ and $\boldsymbol{p}_{d+2}$ can be randomly selected as long as \eqref{eq: march speed requirement} and~\eqref{eq: curvature requirement} are satisfied.
Once the sequence $\{\boldsymbol{p}_d, t_d\}_{d=1}^D$ is obtained, we use it in the minimum-snap algorithm~\citep{mellinger2011minimum} to generate a smooth (piecewise) polynomial trajectory that connects the waypoints sequentially with the traversing time specified as the ToA, i.e., a piecewise polynomial $\mathcal{P}$ of time $t$ that satisfies $\mathcal{P}(t_d) = \boldsymbol{p}_d$ for any $d \in \{1,2,\dots, D\}$.  This procedure is summarized in Algorithm~\ref{algo: traj generation} in the Appendix.
To provide variations in each category, we run Algorithm~\ref{algo: traj generation} several times and collect different polynomials $\mathcal{P}_p(t)$ for $p \in \{1,2,\cdots,P\}$, which we call \textit{parent} polynomials. 

Recall that the sampling period of the system in~\eqref{eq: dynamics} is $t_s$.  Since $t_s \ll \Delta t$, let $\Delta t$ be a multiple of $t_s$. Therefore, over the time window $[t_1,t_D]$ with sampling time $t_s$, one can evaluate a polynomial and obtain $\frac{t_D - t_1}{t_s}+1 = (D-1)\frac{\Delta t}{t_s} + 1$ number of waypoints for the trajectory. However, the length of each task in the trajectory bank must be $M$ because the input to the TPN is defined only for $M$ number of waypoints.  Thus, we have to split these waypoints into consecutive pieces such that each piece will contain $M$ waypoints. Note that we can always choose $D$, $\Delta t$, and $M$ such that $(D-1)\frac{\Delta t}{t_s} + 1$ is a multiple of $M$, i.e., $(D-1)\frac{\Delta t}{t_s} +1 = q M$ where $q$ is a positive integer. 
Mathematically, we can define an $M$-step piece generated by a polynomial as follows:
\begin{minipage}{\linewidth}
    \begin{equation}
    S^{i,j}_{p,s} = \left\{\mathcal P_{p}^{i,j}(t_1 + j t_s), t_1 + j t_s\right\}_{j = sM}^{(s+1)M -1},~~~ s \in \{0,1,\cdots,q-1\},
\end{equation}
\end{minipage}
where $S^{i,j}_{p,s}$ represents the $s$th piece generated by the polynomial of 
parent $p$ under category $\mathcal S_i\mathcal C_j$.  

Finally, we can generate the reference output trajectory, $T^{i,j}_{p,s} = \{\bar{\boldsymbol{y}}^{i,j}_{p,s}(k)\}_{k=1}^M$, based on $\mathcal P_{p}^{i,j}$ and the sequence of sampling time, owing to the quadrotor dynamics' differential flatness~\citep{mellinger2011minimum}. In other words, the $M$-step reference trajectory can be computed by

\noindent
\begin{minipage}{\columnwidth}
\centering
\begin{equation}
    T^{i,j}_{p,s}  = DF\left(\mathcal P_{p}^{i,j}, 
    \{t_1 + j t_s\}_{j = sM}^{(s+1)M -1}
    \right),
\end{equation}
\end{minipage}
where $DF(\cdot)$ refers to the operation using differential flatness that generates a reference output trajectory from the polynomial $\mathcal P_{p}^{i,j}$.  Note that $\bar{\boldsymbol{y}}^{i,j}_{p,s}(k)$ may contain the information on position, velocity, rotation matrix, and angular velocity, for instance, which includes the waypoints in $S^{i,j}_{p,s}$.

\subsection{Batch-DiffTune for Expert Parameters}\label{subsec: batch auto-tuning}

This subsection proposes a controller auto-tuning tool, batch-DiffTune, to obtain the expert control parameter for a parent task $T^{i,j}_{p,s}$ in the bank. Since $T^{i,j}_{p,s}$ includes $M$ steps, the tracking performance can be measured by the cost $L_M(\boldsymbol{\theta};T^{i,j}_{p,s})$ for a given $\boldsymbol{\theta}$. For instance, we can consider the quadratic loss of
the tracking error and control-effort penalty, $L_M(\boldsymbol{\theta};T^{i,j}_{p,s}) = \sum_{k=1}^{M} \norm{\boldsymbol{y}_k -\bar{\boldsymbol{y}}^{i,j}_{p,s}(k)}^2 + \sum_{k=0}^{M-1}\lambda \norm{\boldsymbol{u}_k}^2$ 
with $\lambda>0$ being the penalty coefficient.

Intuitively, one may seek an optimal $\boldsymbol{\theta}^*$ to minimize $L_M(\boldsymbol{\theta};T^{i,j}_{p,s})$, which can be done using the DiffTune algorithm~\citep{cheng2022difftune,cheng2023difftuneplus,tao2023DT-MPC}.
However, DiffTune only focuses on a single task $T^{i,j}_{p,s}$, which may lead to the resulting optimal parameter overfitting to this specific task or a specific initial state $\boldsymbol{x}_0$. To enhance the robustness of the auto-tuned parameter, we propose a batch version of DiffTune, which can produce an expert parameter $\boldsymbol{\theta}^*$ from a batch of tasks.
We randomly perturb the parent's waypoints, $\{\mathcal{P}_p(t_d)\}_{d=1}^D$, to shift within a ball of radius $r$ and generate $C$ child polynomials, $\mathcal{P}_{p,c}(t)$ for $c \in \{1,\cdots,C\}$, with the perturbed waypoints. By doing so, we introduce small variations in speed and curvature of the child polynomials compared with their parent (refer to the illustration shown in Fig.~\ref{fig: generation of trajectory bank}). One can generate the child tasks $\{T^{i,j}_{p,c,s}\}_{c=1}^C$ from polynomial $\mathcal{P}_{p,c}^{i,j}$ in the same approach as their parents $T^{i,j}_{p,s}$ introduced in Section~\ref{sec: task generation}.

Denote the task batch by $\mathcal{B}^{i,j}_{p,s} = \{T^{i,j}_{p,c,s}\}_{c=1}^C$. Since the indices $i,j,p,s$ are fixed for a batch, 
\begin{wrapfigure}{r}{8.7cm}
\centering
\vspace{-4mm}
\fbox{\begin{minipage}[c][3cm]{8.5cm} 
\centering
\vspace{-4mm}
\begin{equation}\label{prob: controller tuning as a parameter optimization}
    \begin{aligned} 
    & \underset{\boldsymbol{\theta} \in \Theta}{\text{min}} && 
    \frac{1}{C} \sum_{c=1}^C L_M\left(\boldsymbol{\theta};T_c\right)\\
    & \text{s.t.} && \boldsymbol{x}_c(k+1) = f\left(\boldsymbol{x}_c(k),\boldsymbol{u}_c(k)\right),\\
    & && \boldsymbol{u}_c(k) = h\left(\boldsymbol{x}_c(0:k),\bar{\boldsymbol{y}}_c(1:k+1),\boldsymbol{\theta}\right), \nonumber \\
    & && \boldsymbol{y}_c(k) = g\left( \boldsymbol{x}_c(k) \right) ,~~c \in \{1,\cdots,C\}.
    \end{aligned}
    \tag{PO}
\end{equation}
\end{minipage}}
\vspace{-8mm}
\end{wrapfigure}
we will drop them and leave $c$ only in the following discussion for notational simplicity if it is clear in context.
For Batch-DiffTune, we find an optimal $\boldsymbol{\theta}^*$ by minimizing the batch loss using the formulation in~\eqref{prob: controller tuning as a parameter optimization}.
Here $\boldsymbol{x}_c(0)$ is randomly generated, satisfying certain distribution (e.g., uniform distribution over a reasonable set of states). 
This practice can eliminate the expert parameters' dependence on the initial state and make the resulting expert parameter depend on tasks only. Subsequently, the parameter update will be conducted by

\noindent
\begin{minipage}{\linewidth}
    \begin{equation}
\label{eq: projected gradient descent to update the parameters}
\boldsymbol{\theta} \leftarrow P_{\Theta}\left(\boldsymbol{\theta} - \frac{\alpha}{C} \sum_{c=1}^C \nabla_{\boldsymbol{\theta}} L_M\left(\boldsymbol{\theta};T_c\right) \right),
\end{equation}
\end{minipage}
where $\mathcal{P}_{\Theta}$ is the projection operator that projects the new parameter value to the feasible set $\Theta$, $\alpha$~is the step size, and $\nabla_{\boldsymbol{\theta}}L_M\left(\boldsymbol{\theta};T_c\right)$ is the gradient of the loss function with respect to parameter $\boldsymbol{\theta}$. 
Batch-DiffTune iteratively unrolls the dynamics and updates parameters in~\eqref{eq: projected gradient descent to update the parameters} until a stop criterion is met (e.g., converging to a minimum or reaching a maximum number of iterations).  Towards the end of the batch gradient descent, we obtain the expert parameter $\boldsymbol{\theta}^*(\mathcal{B})$ for child tasks in $\mathcal{B}$. We collect the dataset containing pairs $\{
T^{i,j}_{p,s}, \boldsymbol{\theta}^*(\mathcal{B}^{i,j}_{p,s})\}$ and use this dataset to train the TPN.

\subsection{Structure and Training of the TPN}

Given the task-parameter dataset $\{
T^{i,j}_{p,s}, \boldsymbol{\theta}^*(\mathcal{B}^{i,j}_{p,s})\}$ obtained in Section~\ref{subsec: batch auto-tuning}, we train the TPN using supervised learning. 
Since the TPN encodes the mapping from tasks to parameters, it can infer parameter choice for unseen tasks (trajectories) whose motion characteristics have already been included in the trajectory bank and, thus, in the training set of the TPN.

We consider the DNN structure for the TPN.
We regulate the last layer of the TPN to output the parameters within the feasible set $\Theta$.  For instance, if $\Theta$ is convex (e.g., a geometric controller's empirical feasible set can be $\Theta = \{\boldsymbol{\theta}|\boldsymbol{\theta} \geq 0\}$), we can use a specially designed last layer~\citep{tordesillas2023rayen} to enforce the hard constraints induced by $\Theta$. 
The last layer is used to ensure that the parameters generated by TPN are within the feasible set $\Theta$ at run time, preserving the stability of the system. 
We use the following mean squared error (MSE) as the loss function for training the TPN:
\begin{minipage}{\linewidth}
    \begin{equation}\label{eq: loss function for training the TPN}
    \text{MSE} = \sum_{i,j,p,s}  \norm{\boldsymbol{\theta}^*(\mathcal{B}^{i,j}_{p,s}) - \phi\left(T^{i,j}_{p,s}\right)}^2.
\end{equation}

\end{minipage}

\section{Simulation Results}\label{sec: sim results}

\noindent \textbf{Generation of the trajectory bank:}
In this case study, we generate 2D trajectories on the horizontal plane covering speeds of $\{1,2,3\}$~m/s (denoted by $\mathcal{S}_1$, $\mathcal{S}_2$, and $\mathcal{S}_3$) and curvature in the range of 0--0.2, 0.2--0.4, 0.4--0.6, 0.6--0.8 (denoted by $\mathcal{C}_1$, $\mathcal{C}_2$, $\mathcal{C}_3$, and $\mathcal{C}_4$). 
Under each category $\mathcal{S}_i \mathcal{C}_j$ for 
\begin{figure}[b]
    {
    \subfigure[][c]
    {
    \includegraphics[width = 0.48\columnwidth]{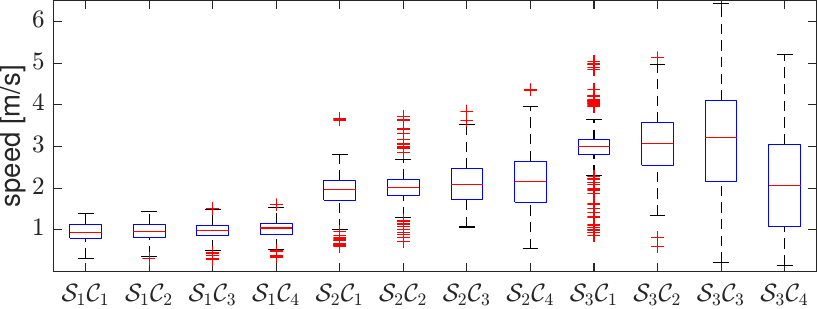}
    }
    \hfill
    \subfigure[][c]
    {
    \includegraphics[width = 0.48\columnwidth]{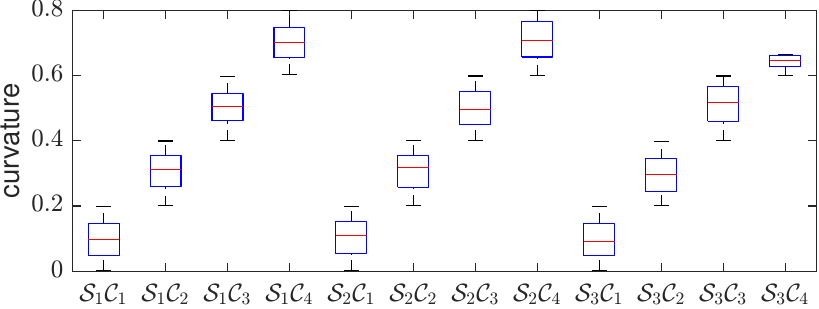} }
    {\caption{Box plots of speed (a) and curvature (b) over the 12 categories in the trajectory bank.  \label{fig:spd and curvature distribution}
    }}
   
    }
    \vspace{-0.3 cm}
\end{figure}
$i \in \{1,2,3\}$ and $j \in \{1,2,3,4\}$, we follow the method in Section~\ref{sec: task generation} and generate 20 trajectories as the parent trajectories. 
Figure~\ref{fig:spd and curvature distribution} shows the speed and curvature distribution among the 20 parent trajectories across the 12 categories. (Figure~\ref{fig:Illustration of all trajectories} in the Appendix shows parent trajectories as an illustration of the trajectory bank.)
One can observe that the trajectories provide consistent speed and curvature in accordance with the associated categories. The only exception is $\mathcal{S}_3 \mathcal{C}_4$, the most aggressive trajectories, where the median speed is smaller than the designated 3 m/s and the curvatures are concentrated towards the lower bound 0.6 of $\mathcal{C}_4$. This phenomenon is anticipated because it is challenging to attain aggressive translational and rotational motions simultaneously.

Each trajectory (both parent and child) contains a total of 15 waypoints (i.e., $D = 14$), where the time increment between adjacent waypoints $\Delta t$ is set to 1 s. 
We exclude the first and last two seconds of the trajectory since they include the transition from/to the initial/terminal waypoint with zero speed. For the remaining 10 s of the trajectory in the middle, {we define each trajectory piece of two-second duration as a \textit{task}}. Hence, each trajectory contains five tasks. 
For each category $\mathcal{S}_i \mathcal{C}_j$, this procedure will produce $20 \text{(parents)} \times 5 \text{(pieces)}=100$ batches of tasks for batch-DiffTune. Overall, the trajectory bank with a total of $3 \times 4=12$ categories can produce $12 \text{(categories)} \times 100 \text{(batches)} =1200$ batches.
The waypoints of the child trajectories are perturbed from those of its parent within a ball of radius $r = 0.05$.
Figure~\ref{fig:Children surround parent} in the Appendix shows example trajectories of parent and child polynomials.

\noindent \textbf{Batch auto-tuning over tasks in the trajectory bank:}
We use the geometric control~\citep{lee2010geometric} with 12 tunable parameters (6 for translational control and 6 for rotational control). Details of the quadrotor dynamics and geometric control are shown in Appendix~\ref{subsec: quadrotor dynamics and control}.
We first 
\begin{wrapfigure}{r}{0.4\textwidth}
    \vspace{-5 mm}

    \includegraphics[width = 0.4\columnwidth]{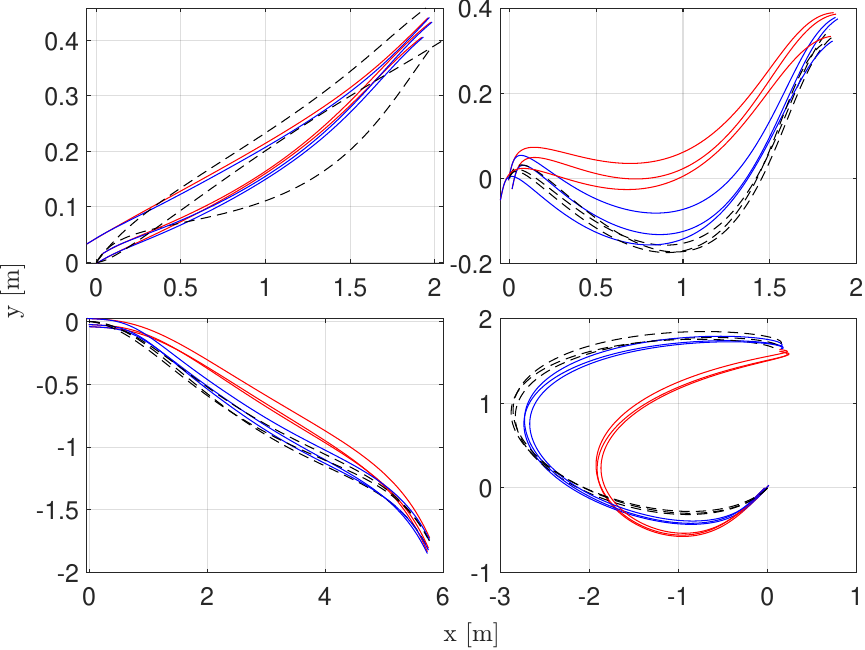}
    \caption{Trajectories ran by expert parameters (blue) vs untrained parameters (red) on three training tasks (black). Only three trajectories are shown for illustration purposes. Subfigures from left to right, top to bottom are from $\mathcal{S}_1 \mathcal{C}_1$, $\mathcal{S}_1 \mathcal{C}_4$, $\mathcal{S}_3 \mathcal{C}_1$, $\mathcal{S}_3 \mathcal{C}_4$.  
    }
    \vspace{-5 mm}
    
    \label{fig: train untrained trajectories}
\end{wrapfigure}
conduct training for the expert parameters over the task batches. Specifically, for any parent task from any category, we use the corresponding child tasks to form a batch to batch-tune one set of parameters (20 child tasks in total: 16 for training and 4 for validation). 
The loss function over each training task is the root-mean-squared error (RMSE) of position tracking. 
We set the step size $\alpha = 0.1$ and run batch-DiffTune for 100 iterations. Overall, the training can effectively reduce the tracking error in all categories. We show the tracking performance before and after training in Fig.~\ref{fig: train untrained trajectories} under the four corner categories ($\mathcal{S}_1 \mathcal{C}_1$: slow and gently curving; $\mathcal{S}_1 \mathcal{C}_4$: slow and sinuous; $\mathcal{S}_3 \mathcal{C}_1$: fast and gently curving; $\mathcal{S}_3 \mathcal{C}_4$: fast and sinuous). The trained parameters are able to steer the quadrotor to closely track the desired trajectories. The trajectory tracking is challenging for the untrained parameters to catch up, especially when the trajectory has a relatively large curvature (e.g., $\mathcal{S}_1 \mathcal{C}_4$ and $\mathcal{S}_3 \mathcal{C}_4$).

\noindent \textbf{Training of the TPN:}
We use the above 1200 pairs ($12 \text{(categories)} \times 20 \text{(parents)} \times 5 \text{(pieces)}$) of task batches and expert parameters to train the TPN. We use a multi-layer perceptron for the TPN. 
The input to the network is a 2D trajectory as a sequence of waypoints on the $(x,y)$-plane (of size $\mathbb{R}^{201 \times 2}$) and reshaped into a vector input of size $\mathbb{R}^{402}$. The output is the expert parameter in the vector form of size $\mathbb{R}^{12}$).
The network has three hidden layers of size [128, 64, 12]. For the last layer, we use RAYEN~\citep{tordesillas2023rayen} to enforce the output of the TPN to be within $\Theta = \{\boldsymbol{\theta} | \boldsymbol{\theta} \geq 0.01\}$, where 0.01 is chosen as the margin to ensure the parameters generated by TPN are positive. We use ReLU as the activation function and apply layer normalization before activation.
The network is trained using an ADAM optimizer with a mean-squared error (MSE) loss function, a learning rate of 0.001, and a batch size of 32. The training is completed in 50 episodes where the MSE error reduces from 88.07 to 0.30 (validation loss reduces from 10.84 to 0.27).

\noindent \textbf{Evaluation of the TPN for flight performance}
We test the TPN's performance when it is used with the controller. We conducted three groups of comparisons to answer the following three questions: 1. How is the performance of TPN compared with expert parameters over the categories in the trajectory bank where the TPN is trained on? 2. How is the performance of TPN compared with expert parameters over categories outside the trajectory bank where the TPN is not trained on? 3. How is the performance of TPN over unseen trajectories of other parametric forms, such as trigonometric trajectories?

To answer the first question, we compare TPN-produced parameters with expert parameters 
\setlength{\tabcolsep}{6pt} 
\renewcommand{\arraystretch}{1} 
  \captionsetup{
	skip=5pt, position = bottom}
\begin{wraptable}{r}{0.75\columnwidth}
	\centering
	\small
	\captionsetup{font=small}
    \vspace{-6mm}
    
 \caption{Comparison of RMSE attained by expert parameters and TPN-produced parameters over the 12 categories in the trajectory bank. Unit: [m]. }
    \label{tab: TPN vs expert in training categories}
    \begin{tabular}{lcccc}
    \toprule[1pt]
         & $\mathcal{C}_1$  & $\mathcal{C}_2$  & $\mathcal{C}_3$  & $\mathcal{C}_4$ \\
         \toprule
         $\mathcal{S}_1$, untrained & 0.203$\pm0.048$	&0.205$\pm0.049$&0.203$\pm0.048$	&0.205$\pm0.051$ \\
        $\mathcal{S}_1$, expert & \textbf{0.180$\pm0.035$} & \textbf{0.176$\pm0.031$}  & \textbf{0.179$\pm0.033$}
  & 0.182$\pm0.035$
 \\
 $\mathcal{S}_1$, TPN & 0.181$\pm0.034$
 & 0.181$\pm0.035$
  & 0.181$\pm0.034$
 & \textbf{0.179$\pm0.034$} \\ 
 \midrule
$\mathcal{S}_2$, untrained &  0.212$\pm0.056$ & 0.290$\pm0.088$ &	0.285$\pm0.082$	& 0.378$\pm0.088$\\
        $\mathcal{S}_2$, expert & \textbf{0.180$\pm0.035$}
 & \textbf{0.184$\pm0.041$}
 & \textbf{0.182$\pm0.040$}
 & \textbf{0.191$\pm0.048$}
 \\
 $\mathcal{S}_2$, TPN & 0.180$\pm0.035$
 & 0.187$\pm0.043$ & 0.188$\pm0.044$
 & 0.191$\pm0.050$ \\ 
\midrule
$\mathcal{S}_3$, untrained & 0.241$\pm0.072$ &	0.311$\pm0.087$ &	0.656$\pm0.122$ &	0.868$\pm0.100$ \\
        $\mathcal{S}_3$, expert & \textbf{0.181$\pm0.037$}
 & \textbf{0.185$\pm0.042$}
 & \textbf{0.208$\pm0.062$}
 & \textbf{0.223$\pm0.071$}
\\
$\mathcal{S}_3$, TPN & 0.183$\pm0.039$ &0.194$\pm0.048$
 & 0.214$\pm0.065$
 & 0.228$\pm0.073$  \\ 
 
 \bottomrule[1pt]
    \end{tabular}
        \vspace{-6mm}
\end{wraptable}
over the tasks in the 12 categories used in the TPN's training. {To evaluate the performance over each category, we test each set of parameters over trajectory piece 1 of parents 17-20. In addition, for each parent trajectory, the performance is evaluated with simulations including randomized initial state across 16 combinations, each involving a positional and velocity shift of 0.3 m and 0.3 m/s, respectively, in any direction along the $x$- and $y$-axes.} Table~\ref{tab: TPN vs expert in training categories} shows the mean and standard deviation of tracking RMSEs {from the 16 combinations of initial states.}
As can be seen, through all 12 categories, TPN produces almost identical performance to the expert, {with a maximum degradation of 4.9\% in $\mathcal{S}_3 \mathcal{C}_2$}. This result shows the consistent performance that TPN can produce compared to the expert parameters.

To answer the second question, we compare TPN-produced parameters with expert parameters over the tasks in 8 categories excluded by the trajectory bank and subsequently not used in the TPN's 
\setlength{\tabcolsep}{6pt} 
\renewcommand{\arraystretch}{1} 
  \captionsetup{
	skip=5pt, position = bottom}
\begin{wraptable}{r}{0.55\columnwidth}
	\centering
	\small
	
	\vspace{-6mm}
	\captionsetup{font=small}
     \caption{Comparison of RMSE attained by expert parameters, TPN parameters, and untrained parameters over unseen tasks in the categories that are not in the trajectory bank. Unit: [m].}
    \label{tab: OOD testing with minsnap trajectories}
    \begin{tabular}{cccc}
    \toprule[1pt]
         & Expert & TPN & Untrained \\
         \midrule
        $\mathcal{S}_1 \mathcal{C}_5$ &0.182$\pm0.037$ & 
        \textbf{0.180$\pm0.036$} 
        & 0.215$\pm0.058$  \\
        $\mathcal{S}_1 \mathcal{C}_6$ & \textbf{0.178$\pm0.035$} & 
        0.180$\pm0.037$
        & 0.223$\pm0.064$ \\
        $\mathcal{S}_2 \mathcal{C}_5$ & 0.208$\pm0.061$ & 
        \textbf{0.189$\pm0.055$}  
        & 0.480$\pm0.192$ \\
        $\mathcal{S}_4 \mathcal{C}_1$ & \textbf{0.192$\pm0.049$} & 
        0.204$\pm0.055$
        & 0.313$\pm0.129$\\
        $\mathcal{S}_4 \mathcal{C}_2$ & \textbf{0.206$\pm0.061$} & 
        0.236$\pm0.069$  
        & 0.487$\pm0.199$\\
        $\mathcal{S}_4 \mathcal{C}_3$ & \textbf{0.232$\pm0.077$} & 
        0.273$\pm0.084$ 
        & 0.882$\pm0.373$ \\
        $\mathcal{S}_5 \mathcal{C}_1$ & \textbf{0.208$\pm0.062$} & 
        0.244$\pm0.071$ 
        & 0.485$\pm0.194$\\
        $\mathcal{S}_6 \mathcal{C}_1$ & \textbf{0.207$\pm0.063$} & 
        0.263$\pm0.086$ 
        & 0.512$\pm0.285$ \\ \bottomrule[1pt]
    \end{tabular}

   \vspace{-4mm}
\end{wraptable}
training (shown in Table~\ref{tab: OOD testing with minsnap trajectories}). These categories either have speeds bigger than 3~m/s (i.e., $\mathcal{S}_4$, $\mathcal{S}_5$, and $\mathcal{S}_6$) or have curvatures that go beyond the training categories (i.e., $\mathcal{C}_5$ and $\mathcal{C}_6$ for curvatures in the range of 0.8--1 and 1--1.2). {We apply the same initial states (16 combinations of position and velocity shifts) described in the first test}. The results are shown in Table~\ref{tab: OOD testing with minsnap trajectories}, in which the TPN's performance is suboptimal to that of the experts trained for these categories. When including the untrained parameters in this comparison, the untrained parameters show much worse tracking performance than both the expert and TPN parameters. The overall comparison suggests that the TPN has gained generalization capability to the tasks not seen in the training of TPN.

To answer the third question, we introduce trajectories characterized by trigonometric functions that 
are completely different from the piecewise polynomial trajectories during training. 
In 
\setlength{\tabcolsep}{6pt} 
\renewcommand{\arraystretch}{1} 
  \captionsetup{
	skip=5pt, position = bottom}
\begin{wraptable}{r}{0.4\columnwidth}
	\centering
	\small
	
	\vspace{-6mm}
	\captionsetup{font=small}
     \caption{Performance comparison between TPN parameters and untrained parameters, over trajectories parameterized by trigonometric functions. Unit: [m].}
    \label{tab: OOD testing with trigonometric functions}
    \begin{tabular}{cccc}
    \toprule[1pt]
          & TPN & Untrained \\
          \midrule
        Cir(1)  & 
        \textbf{0.132$\pm0.026$} 
        & 0.149$\pm0.035$ \\
        Cir(2)  &
        \textbf{0.136$\pm0.033$}  
& 0.226$\pm0.056$ \\
        Cir(3)  & 
        \textbf{0.141$\pm0.039$}  
        & 0.461$\pm0.068$ \\
        Cir(4)  & 
        \textbf{0.211$\pm0.056$} 
        & 0.861$\pm0.071$\\
        Lem(1)  & 
        \textbf{0.131$\pm0.025$} 
        & 0.143$\pm0.033$\\
        Lem(2)  & 
        \textbf{0.130$\pm0.025$} 
        & 0.143$\pm0.033$ \\
        Lem(3)  & 
        \textbf{0.129$\pm0.025$} 
        & 0.145$\pm0.033$ \\
        Lem(4)  & 
        \textbf{0.125$\pm0.026$}  
        & 0.153$\pm0.037$ \\
        \bottomrule[1pt]
    \end{tabular}
\end{wraptable}
this scenario, we select a circular trajectory (denoted by Cir($v$) for $x(t) = 1-\cos(vt)$ m and $y(t) = \sin(vt)$ m; $v\in \mathbb{R}$ is the linear speed) and a Leminiscate trajectory (denoted by Lem($v$) for $x(t) = \sin(2vt/2.5)$ m and $y(t) = 1.5\sin(vt/2.5)$ m; $v \in \mathbb{R}$ is the maximum speed on the trajectory). Since we do not train experts in these trajectories, we compare TPN-produced parameters with untrained parameters. We apply the same initial states (16 combinations of position and velocity shifts) described in the first test. The results are shown in 
Table~\ref{tab: OOD testing with trigonometric functions}. Even though these trigonometric functions were not seen when training for TPN, the relationship between the geometric characteristics and desirable parameters is captured by the TPN, which leads to improved and more consistent performance than the untrained ones. The latter's performance deteriorates subject to the increased speed of the trigonometric trajectories. 

\section{Conclusions}\label{sec: conclusion}

This paper addresses the problem of determining an appropriate set of control parameters for unforeseen tasks at runtime using the TPN, where the TPN encodes the mapping from tasks to control parameters. 
We construct a trajectory bank that contains tasks subject to a range of motion characteristics and obtain the associated expert parameters by auto-tuning the MBC over the tasks from the trajectory bank. We train the TPN using pairs of tasks and expert parameters by supervised learning.
The TPN achieves almost expert-level performance on the unseen tasks within the trajectory bank, suboptimal performance on the tasks within unseen categories that are outside the trajectory bank, and consistent performance over unseen tasks parameterized differently than those in the trajectory bank. For future work, we will demonstrate the capability of a TPN on a real quadrotor system and provide theoretical guarantees over the system properties when using the TPN. Extension of the current setup to other types of systems will also be considered, such as legged robots.

\acks{This work is supported by NASA Cooperative Agreement (80NSSC20M0229), NASA ULI (80NSSC22M0070), AFOSR (FA955021-1-0411), AFOSR DURIP (FA9550-23-1-0129), NSF M3X (2431216), NSF-AoF Robust Intelligence (2133656), and NSF SLES (2331878, 2331879).
}

\bibliography{ref}

\section*{Appendix}
\subsection{Trajectory Generation for the Trajectory Bank}
\begin{wrapfigure}{r}{0.5\columnwidth}
  \begin{minipage}{\linewidth}
\setlength{\textfloatsep}{1pt}
\vspace{-1cm}

\begin{algorithm}[H]
	\caption{Trajectory generation for the trajectory bank }
	\label{algo: traj generation}

	\begin{algorithmic}[1]
		\REQUIRE Time increment $\Delta t$, total number of waypoints $D$ in one trajectory, designated speed $v$, designated range of curvature $\mathcal{C}$.
		\ENSURE A smooth trajectory within the category of speed $v$ and curvature $\mathcal{C}$.
		\STATE{Initialize waypoints $\boldsymbol{p}_1$ at the origin and $\boldsymbol{p}_2$ at ($v \Delta t$,0)}
            \STATE{Initialize $t_1 \leftarrow 0$ and $t_2 \leftarrow \Delta t$.}
            \FOR{$d \gets 3$ to $D$}    
		\REPEAT 
            \STATE{Sample a candidate waypoint $\tilde{\boldsymbol{p}}_d$ on the ball centered on $\boldsymbol{p}_{d-1}$ with radius $v  \Delta t$.}
            \STATE{Compute Menger curvature~\citep{menger1930untersuchungen} $c=1/R(\boldsymbol{p}_{d-2}, \boldsymbol{p}_{d-1}, \tilde{\boldsymbol{p}}_{d})$.}
        \UNTIL{$c \in \mathcal{C}$}
        \STATE{update the new waypoint $\boldsymbol{p}_d \leftarrow \tilde{\boldsymbol{p}}_d$ and $t_d \leftarrow (d-1)\Delta t$}
        \ENDFOR
        \STATE {Generate a minimum-snap trajectory~\citep{mellinger2011minimum} with time-stamped waypoints $\{\boldsymbol{p}_d, t_d\}_{d=1}^{D}$}
        \RETURN{the minimum-snap trajectory}
	\end{algorithmic}
\end{algorithm}
\vspace{-1.5cm}
  \end{minipage}
\end{wrapfigure}

The algorithm is summarized in Algorithm~\ref{algo: traj generation}.

\begin{remark}
    We choose to use smooth (piecewise) polynomials to construct the trajectory bank since any smooth trajectory in the space of flat outputs (3D position and yaw angle) can be tracked by  quadrotors~\citep{mellinger2011minimum}. Besides, smooth (piecewise) polynomials are widely adopted in upper-stream planning modules~\citep{gao2018online,zhou2019robust}, which can extend the trajectory bank to include planning-oriented tasks. Note that trigonometric functions are also smooth functions for quadrotor trajectory tracking tasks, especially for benchmarking controller performance~\citep{{sun2022comparative,saviolo2022physics,wu20221},wu2023L1Quad}.
    However, they uniformly approximate  \textit{periodic} continuous functions only, which limits the types of tasks for quadrotors in practice and hence they are not applied in this work.
\end{remark}

\begin{remark}
    The TPN's input is a task in a non-parametric form of time-stamped reference points, which is the most general form, incorporating trajectories that are parametrized differently. Furthermore, the reference points contain richer information than just position information: Consecutive two/three/four/five reference points in the time series encode the velocity/acceleration/jerk/snap information of the trajectory through finite-difference approximations. The rich information encoded in the time series can be extracted as features by the TPN to facilitate better encoding of the tasks' characteristics for better inference of suitable parameters. 
\end{remark}

\subsection{Illustrations of Trajectories in the Trajectory Bank}
\begin{figure}
    \centering
    \includegraphics[width = \textwidth]{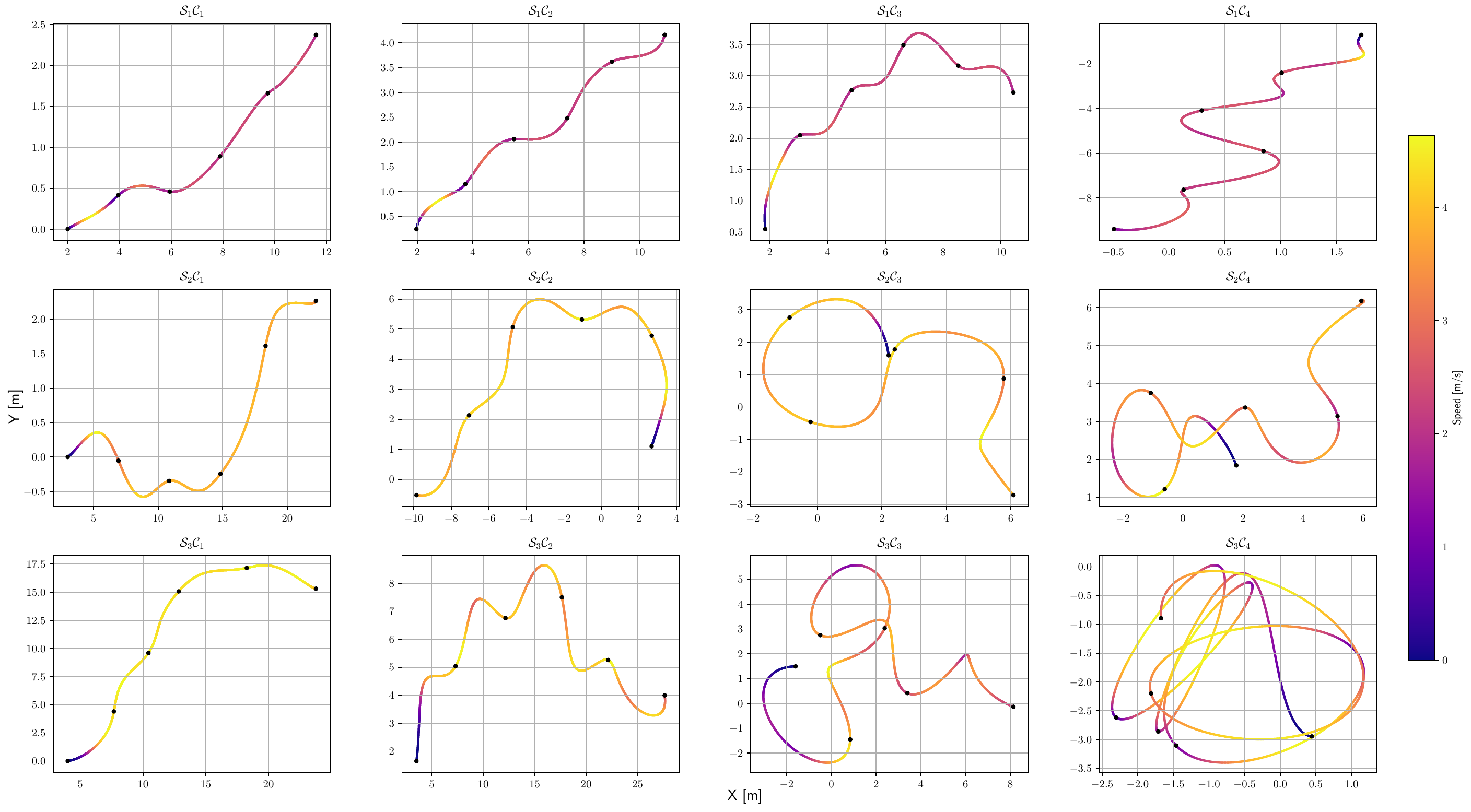} 
    \caption{Illustration of parent~1 over all 12 categories. Each segment in between two dots has a duration of 2 s and represents one \textit{task}. 
    With increased speeds (from $\mathcal{S}_1$ to $\mathcal{S}_3$), the waypoints (dots) are wider spread, corresponding to faster speeds that pose challenges to translational motion tracking. With increased curvatures (from $\mathcal{C}_1$ to $\mathcal{C}_4$), the trajectories turn curvy and sinuous, corresponding to faster turns that pose challenges to rotational motion tracking.}
    \label{fig:Illustration of all trajectories}
\end{figure}

\begin{figure}[h]
    \centering
    \includegraphics[width = 0.5\columnwidth]{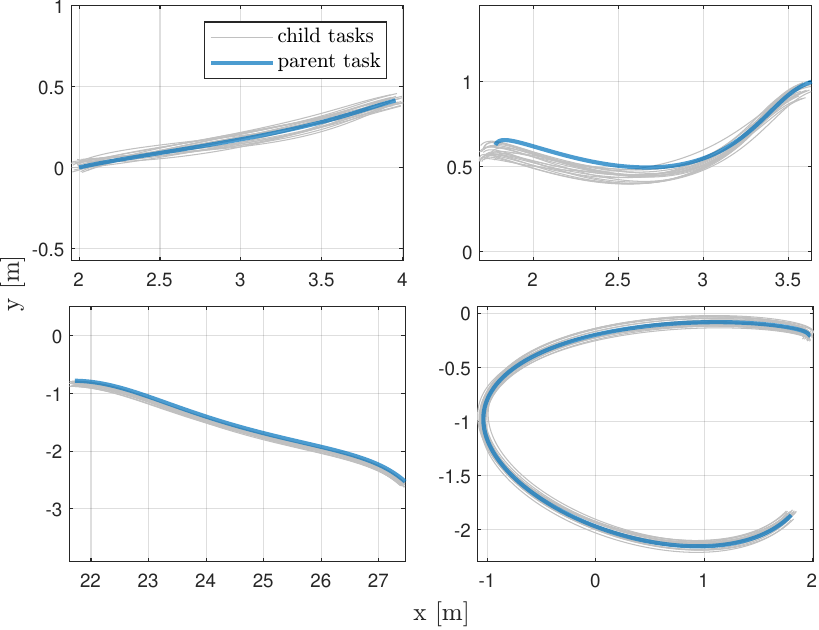}
    \caption{Illustration of parent trajectory and randomized child trajectories surrounding the parent. Subfigures from left to right, top to bottom are from $\mathcal{S}_1 \mathcal{C}_1$, $\mathcal{S}_1 \mathcal{C}_4$, $\mathcal{S}_3 \mathcal{C}_1$, $\mathcal{S}_3 \mathcal{C}_4$.}
    \label{fig:Children surround parent}
\end{figure}

We show the parent trajectory~1 over all 12 categories in Fig.~\ref{fig:Illustration of all trajectories}. The randomized child tasks are shown in Fig.~\ref{fig:Children surround parent} in reference to the parent task.

\subsection{Parameter quality by Batch-DiffTune}
\begin{wrapfigure}{r}{0.5\textwidth}
    \includegraphics[width = 0.5\columnwidth]{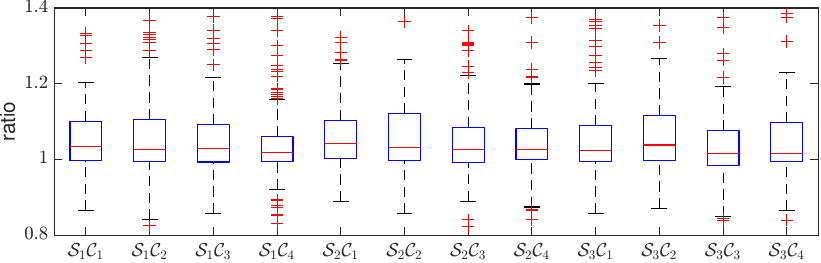}
    \caption{Evaluation of the expert parameters in the testing tasks shown in a box plot. The vertical axis refers to the ratio of mean RMSE achieved by expert parameters on the training tasks divided by those on the testing tasks. 
    }
    \label{fig: train test ratio}
   \vspace{-5mm}
\end{wrapfigure}
To validate batch-tuned expert parameters, we 
evaluate these parameters over the testing set. Recall 
that 16 tasks in each batch were used in batch-DiffTune for auto-tuning. We set the remaining 4 tasks in each batch as the testing set. We divide the average RMSE on the training set by the
average RMSE on the testing set using the parents trained for this batch. We then group these parameters by category and show the statistics of the RMSE ratios in Fig.~\ref{fig: train test ratio}. The ratios have a consistent median for all groups, indicating the trained parameters have achieved a similar level of performance over the testing set compared with their performance on the training set.

\subsection{Quadrotor Dynamics and Geometric Control~\citep{lee2010geometric}}\label{subsec: quadrotor dynamics and control}
\allowdisplaybreaks[0] 
Consider the following nonlinear dynamics model for a quadrotor:

\begin{equation}\label{equ:quadrotor dynamics}
\begin{aligned}
    & \dot{\boldsymbol{p}}=  \boldsymbol{v},\quad &&\dot{\boldsymbol{v}}=  g\boldsymbol{e}_3-\frac{f}{m}R\boldsymbol{e}_3,\\
    & \dot{R} =  R \boldsymbol{\Omega}^{\times},\quad &&\dot{\boldsymbol{\Omega}}=  J^{-1}(\boldsymbol{M} - \boldsymbol{\Omega} \times J \boldsymbol{\Omega}) ,
\end{aligned}
\end{equation}
where $\boldsymbol{p} \in \mathbb{R}^3$ and $\boldsymbol{v} \in \mathbb{R}^3$ are the position and velocity of the quadrotor, respectively, $R \in SO(3)$ is the rotation matrix describing the quadrotor's attitude, ${\boldsymbol{\Omega} \in \mathbb{R}^3}$ is the angular velocity, $g$ is the gravitational acceleration, $m$ is the vehicle mass, $J \in \mathbb{R}^{3 \times 3}$ is the moment of inertia (MoI) matrix, $f$ is the collective thrust, and $\boldsymbol{M} \in \mathbb{R}^3$ is the moment applied to the vehicle. Both the inertial frame and body frame use a North-East-Down (NED) coordinate system.
The \textit{wedge} operator $\cdot^{\times}:\mathbb{R}^3 \rightarrow \mathfrak{so}(3)$ denotes the mapping to the space of skew-symmetric matrices. The control actions $f$ and $\boldsymbol{M}$ are computed using the geometric controller~\citep{lee2010geometric}. We use mass $m = 4.34$ kg and inertia $J = \text{diag}(0.082,0.0845,0.1377)$ kg$\cdot$m${}^2$ for the quadrotor in the simulation, where the dynamics \eqref{equ:quadrotor dynamics} is discretized to 0.01 s.

The design of the geometric controller follows \citep{lee2010geometric,mellinger2011minimum}, where the goal is to have the quadrotor follow prescribed trajectory $\boldsymbol{p}(t) \in \mathbb{R}^3$ and yaw $\boldsymbol{\psi}(t)$ for time $t$ in a prescribed interval $[0,t_f]$. Towards this end, the translational motion is controlled by the desired thrust $f_b$:
\begin{equation}\label{eq:thrust control}
    f = -\bar{\boldsymbol{F}} \cdot (Re_3),
\end{equation}
which is obtained by projecting the desired force $F_d$ to the body-fixed $z$-axis $Re_3$. The desired force $\bar{\boldsymbol{F}} \in \mathbb{R}^3$ is computed via $\bar{\boldsymbol{F}} = -\boldsymbol{k}_{\boldsymbol{p}} \boldsymbol{e}_{\boldsymbol{p}}-\boldsymbol{k}_{\boldsymbol{v}} \boldsymbol{e}_{\boldsymbol{v}} - mg \boldsymbol{e}_3 + m \ddot{\bar{\boldsymbol{p}}}$ for $\boldsymbol{k}_{\boldsymbol{p}} $ and $\boldsymbol{k}_{\boldsymbol{v}}  \in \mathbb{R}^{3}$ being user-selected positive control parameters and $\boldsymbol{e}_{\boldsymbol{p}} = \boldsymbol{p}-\bar{\boldsymbol{p}}$ and $\boldsymbol{e}_{\boldsymbol{v}} = \boldsymbol{v}-\dot{\bar{\boldsymbol{p}}}$ standing for the position and velocity errors, respectively.
The rotational motion is controlled by the desired moment
\begin{align}\label{eq:torque control}
    \boldsymbol{M} = -\boldsymbol{k}_R \boldsymbol{e}_R - \boldsymbol{k}_{\boldsymbol{\Omega}} \boldsymbol{e}_{\boldsymbol{\Omega}} + \boldsymbol{\Omega} \times J \boldsymbol{\Omega} - J( \boldsymbol{\Omega}^{\wedge} R^\top \bar{R} \bar{\Omega}-R^\top \bar{R} \dot{\bar{\boldsymbol{\Omega}}}),
\end{align}
where $\boldsymbol{k}_R$ and $\boldsymbol{k}_{\boldsymbol{\Omega}} \in \mathbb{R}^{3}$ are user-selected positive control parameters; $\bar{R}$, $\bar{\Omega}$, and $\dot{\bar{\Omega}}$ are the desired rotation matrix, desired angular velocity, and desired angular velocity derivative, respectively; $\boldsymbol{e}_R = ( \bar{R}^\top R - R^\top \bar{R})^{\vee}/2$ and $\boldsymbol{e}_{\boldsymbol{\Omega}} = \boldsymbol{\Omega} -R^\top \bar{R} \bar{\boldsymbol{\Omega}}$ are the rotation error and angular velocity error, respectively. The derivations of $\bar{R}$, $\bar{\boldsymbol{\Omega}}$, and $\dot{\bar{\boldsymbol{\Omega}}}$ are omitted for simplicity; see~\citep{lee2010geometricarxiv} for details. 

The geometric controller has a 12-dimensional parameter space, which splits into four groups of parameters: $\boldsymbol{k}_{\boldsymbol{p}}$, $\boldsymbol{k}_{\boldsymbol{v}}$, $\boldsymbol{k}_R$, $\boldsymbol{k}_{\boldsymbol{\Omega}}$ (applying to the tracking errors in position, linear velocity, attitude, and angular velocity, respectively).
Each group is a 3-dimensional vector (associated with the $x$-, $y$-, and $z$-component in each's corresponding tracking error). 
The initial parameters for auto-tuning are set as $\boldsymbol{k}_{\boldsymbol{p}} = 16 \mathbb{I}$, $\boldsymbol{k}_{\boldsymbol{v}} = 5.6\mathbb{I}$, $\boldsymbol{k}_R = 8.81\mathbb{I}$, and $\boldsymbol{k}_{\boldsymbol{\Omega}} = 2.54\mathbb{I}$, for $\mathbb{I} = [1,1,1]^\top$. 

\end{document}